\documentclass[11pt,a4paper]{article}
\usepackage[nohyperref]{acl2018} 
\usepackage{times}
\usepackage{latexsym}
\usepackage{graphicx}
\usepackage{adjustbox}
\usepackage{subfigure}
\usepackage{amsfonts}
\usepackage{url}
\usepackage{color}
\usepackage{multirow}

\usepackage{todonotes} 

\aclfinalcopy 
\def\aclpaperid{1054} 

\DeclareGraphicsExtensions{.pdf,.png,.jpg}

\newcommand\DR{SAN }

\title{Stochastic Answer Networks for Machine Reading Comprehension}

\author{Xiaodong Liu$^\bold{\dagger}$, Yelong Shen$^\bold{\dagger}$, Kevin Duh$^\bold{\ddagger}$ and Jianfeng Gao$^\bold{\dagger}$ \\
  $^\bold{\dagger}$    
  Microsoft Research, Redmond, WA, USA \\
  $^\bold{\ddagger}$
  Johns Hopkins University, Baltimore, MD, USA \\
  {\tt $^\bold{\dagger}$\{xiaodl,yeshen,jfgao\}@microsoft.com
   $^\bold{\ddagger}$kevinduh@cs.jhu.edu}
}

\date{}

\begin{document}
\maketitle
\begin{abstract}
We propose a simple yet robust \textbf{s}tochastic \textbf{a}nswer \textbf{n}etwork (SAN) that simulates multi-step reasoning in machine reading comprehension. 
Compared to previous work such as ReasoNet which used reinforcement learning to determine the number of steps, the unique feature is the use of a kind of stochastic prediction dropout on the answer module (final layer) of the neural network during the training. 
We show that this simple trick improves robustness and achieves results competitive to the state-of-the-art on the Stanford Question Answering Dataset (SQuAD), the Adversarial SQuAD, and the Microsoft MAchine Reading COmprehension Dataset (MS MARCO).
\end{abstract}

\section{Introduction}
\label{sec:intro}
Machine reading comprehension (MRC) is a challenging task: the goal is to have machines read a text passage and then answer any question about the passage.
This task is an useful benchmark to demonstrate natural language understanding, and also has important applications in e.g.~conversational agents and customer service support.
It has been hypothesized that difficult MRC problems require some form of multi-step synthesis and reasoning. 
For instance, the following example from the MRC dataset SQuAD \cite{rajpurkar2016squad} illustrates the need for synthesis of information across sentences and multiple steps of reasoning:\\[0.2cm]
$Q$: What collection does \textbf{the V\&A Theator \& Performance galleries} hold?\\[0.2cm]
$P$: \textbf{The V\&A Theator \& Performance galleries} opened in March 2009. ... \textbf{They} hold the UK's biggest national collection of \underline{material about live performance.}\\[0.2cm]
To infer the answer (the underlined portion of the passage $P$), the model needs to first perform coreference resolution so that it knows ``\textbf{They}'' refers ``\textbf{V\&A Theator}'', then extract the subspan in the direct object corresponding to the answer. 

This kind of iterative process can be viewed as a form of multi-step reasoning. 
Several recent MRC models have embraced this kind of multi-step strategy, where 
predictions are generated after making multiple passes through the same text and integrating intermediate information in the process.
The first models employed a predetermined fixed number of steps \cite{hill2015goldilocks,dhingra2016gated,sordoni2016iterative,kumar15askme}.
Later, \newcite{shen2016reasonet} proposed using reinforcement learning to dynamically determine the number of steps based on the complexity of the question.
Further, \newcite{shen2017empirical} empirically showed that dynamic multi-step reasoning outperforms fixed multi-step reasoning, 
which in turn outperforms single-step reasoning on two distinct MRC datasets (SQuAD and MS MARCO). 

\begin{figure}[t]
    \centering
\adjustbox{trim={.01\width} {.1\height} {.05\width} {.02\height},clip}{\includegraphics[scale=0.75]{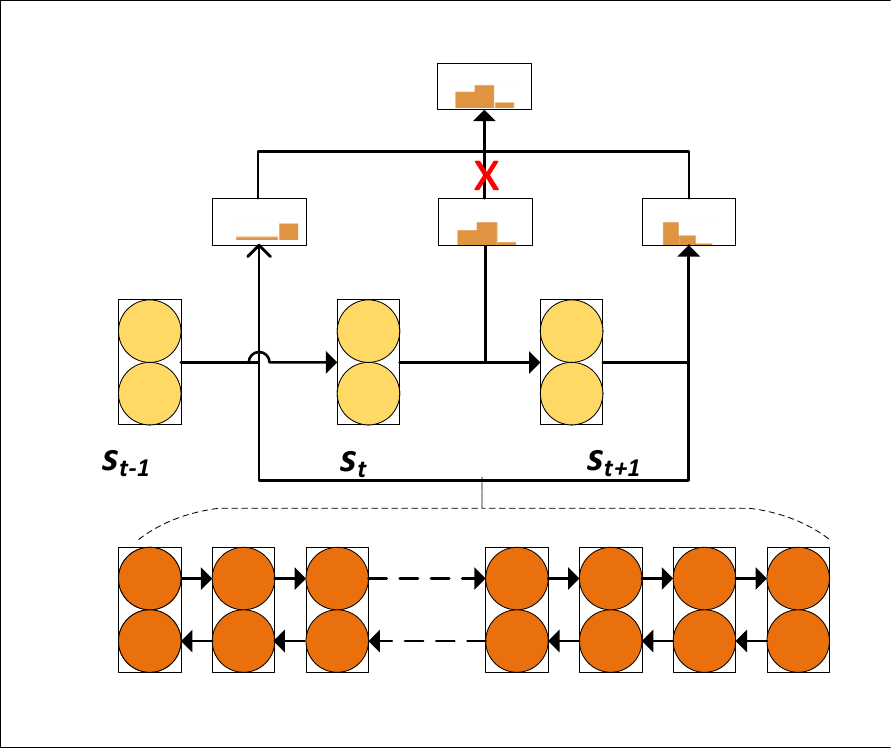}}
\caption{\label{fig:answer-module} Illustration of ``stochastic prediction dropout" in the answer module during training. At each reasoning step $t$, the model combines memory (bottom row) with hidden states $\mathbf{s_{t-1}}$ to generate a prediction (multinomial distribution). Here, there are three steps and three predictions, but one prediction is dropped and the final result is an average of the remaining distributions.
} 
\end{figure}  
  
%

In this work, we derive an alternative multi-step reasoning neural network for MRC. 
During training, we fix the number of reasoning steps, but perform stochastic dropout on the answer module (final layer predictions).
During decoding, we generate answers based on the average of predictions in all steps, rather than the final step. 
We call this a stochastic answer network (SAN) because the stochastic dropout is applied to the answer module; 
albeit simple, this technique significantly improves the robustness and overall accuracy of the model. 
Intuitively this works because while the model successively refines its prediction over multiple steps, each step is still trained to generate the same answer;
we are performing a kind of stochastic ensemble over the model's successive prediction refinements. 
Stochastic prediction dropout is illustrated in Figure \ref{fig:answer-module}.

\section{Proposed model: SAN}
\label{sec:model}

\begin{figure*}[h!]
\centering
\adjustbox{trim={.01\width} {.1\height} {.05\width} {.02\height},clip}%
  {\includegraphics[scale=0.65]{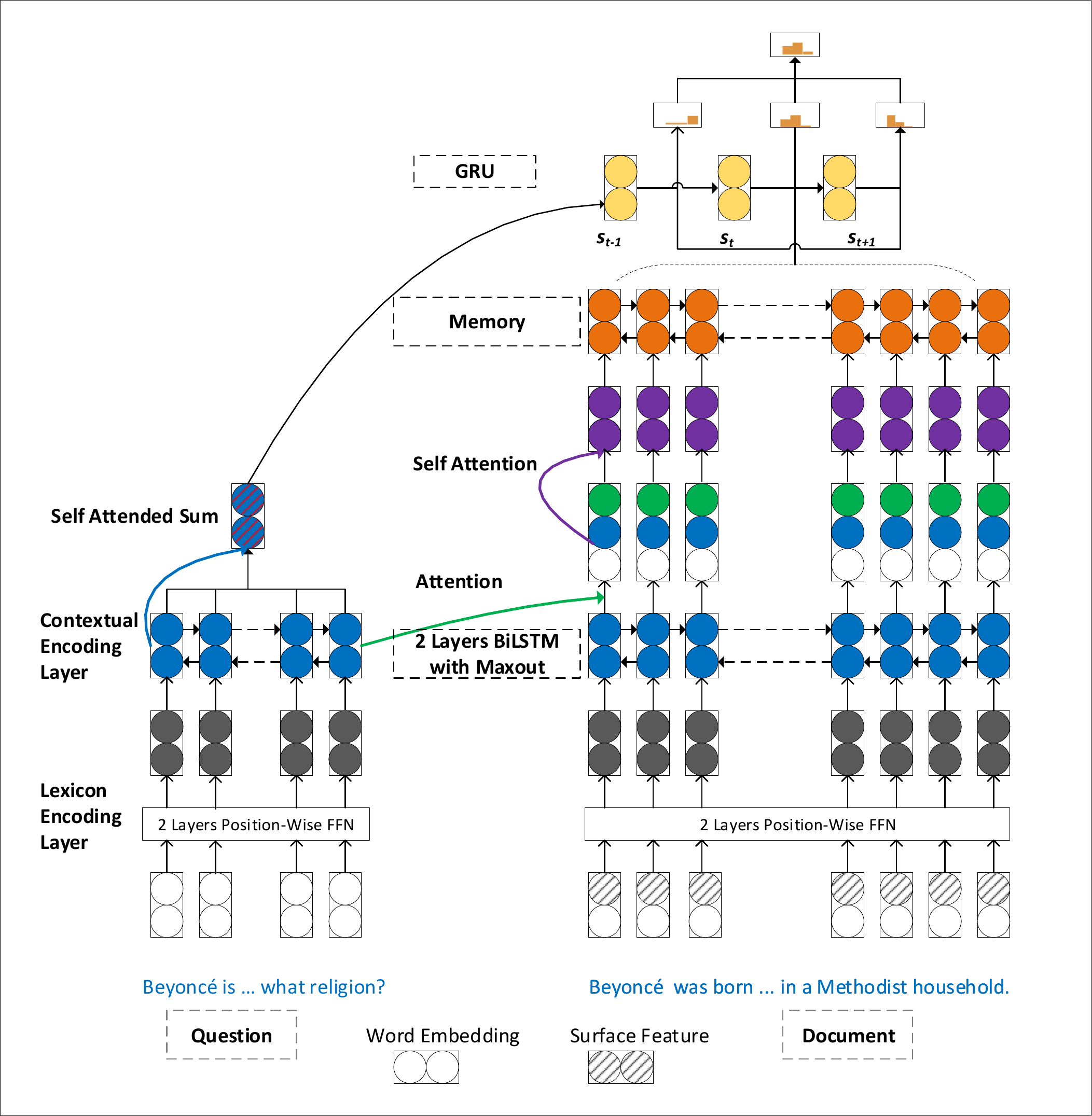}}
\caption{\label{fig:model} {\bf Architecture of the \DR for Reading Comprehension:} The first layer is a lexicon encoding layer that maps words to their embeddings independently for the question (left) and the passage (right): this is a concatenation of word embeddings, POS embeddings, etc. followed by a position-wise FFN. The next layer is a context encoding layer, where a BiLSTM is used on the top of the lexicon embedding layer to obtain the context representation for both question and passage. In order to reduce the parameters, a maxout layer is applied on the output of BiLSTM. The third layer is the working memory: First we compute an alignment matrix between the question and passage using an attention mechanism, and use this to derive a question-aware passage representation. Then we concatenate this with the context representation of passage and the word embedding, and employ a self attention layer to re-arrange the information gathered. Finally, we use another LSTM to generate a working memory for the passage. At last, the fourth layer is the answer module, which is a GRU that outputs predictions at each state $s_t$.}
\end{figure*}

The machine reading comprehension (MRC) task as defined here involves a question $Q=\{q_0, q_1, ..., q_{m-1}\}$ and a passage $P=\{p_0, p_1, ..., p_{n-1}\}$ 
and aims to find an answer span $A=\{a_{start}, a_{end}\}$ in $P$. 
We assume that the answer exists in the passage $P$ as a contiguous text string. 
Here, $m$ and $n$ denote the number of tokens in $Q$ and $P$, respectively. 
The learning algorithm for reading comprehension is to learn a function $f(Q, P) \rightarrow A$. 
The training data is a set of the query, passage and answer tuples $<Q, P, A>$. 

We now describe our model from the ground up. The main contribution of this work is the answer module, but in order to understand what goes into this module, we will start by describing how $Q$ and $P$ are processed by the lower layers. Note the lower layers also have some novel variations that are not used in previous work. 
As shown in Figure \ref{fig:model}, our model contains four different layers to capture different concept of representations.
The detailed description of our model is provided as follows.

\textbf{Lexicon Encoding Layer}. 
The purpose of the first layer is to extract information from $Q$ and $P$ at the word level and normalize for lexical variants.
A typical technique to obtain lexicon embedding is concatenation of its word embedding with other linguistic embedding such as those derived from Part-Of-Speech (POS) tags. 
For word embeddings, we use the pre-trained 300-dimensional GloVe vectors \cite{pennington2014glove} for the both $Q$ and $P$.
Following \newcite{chen2017reading}, we use three additional types of linguistic features for each token $p_i$ in the \textit{passage} $P$:
\begin{itemize}
\item $9$-dimensional POS tagging embedding for total 56 different types of the POS tags.\vspace{-0.2cm}
\item $8$-dimensional named-entity recognizer (NER) embedding for total 18 different types of the NER tags. We utilized small embedding sizes for POS and NER to reduce model size. They mainly serve the role of coarse-grained word clusters.
\item A $3$-dimensional binary \textit{exact} match feature defined as $f_{exact\_match}(p_i)= \mathbb{I}(p_i \in Q)$. This checks whether a passage token $p_i$ matches the original, lowercase or lemma form of any question token.
\item Question enhanced passages word embeddings: $f_{align}(p_i)=\sum_{j}\gamma_{i,j}g(GloVe(q_j))$, where $g(\cdot)$ is a 280-dimensional single layer neural network $ReLU(W_0x)$ and $\gamma_{i,j}=\frac{exp(g(GloVe(p_j))\cdot g(GloVe(q_i)))}{\sum_{j'}exp(g(GloVe(p_i))\cdot g(GloVe(q_{j'})))}$ measures the similarity in word embedding space between a token $p_i$ in the passage and a token $q_j$ in the question. Compared to the \textit{exact} matching features, these embeddings encode \textit{soft} alignments between similar but not-identical words.

\end{itemize}
In summary, each token $p_i$ in the passage is represented as a 600-dimensional vector and each token $q_j$ is represented as a 300-dimensional vector. 

Due to different dimensions for the passages and questions, in the next layer two different bidirectional LSTM (BiLSTM) \cite{hochreiter1997long} may be required to encode the contextual information. 
This, however, introduces a large number of parameters.
To prevent this, we employ an idea inspired by \cite{vaswani2017attention}: use two separate two-layer position-wise Feed-Forward Networks (FFN), $FFN(x)=W_2 ReLU(W_1 x +b_1) + b_2$, to map both the passage and question lexical encodings into the same number of dimensions. Note that this FFN has fewer parameters compared to a BiLSTM.
Thus, we obtain the final lexicon embeddings for the tokens in $Q$ as a matrix $E^q \in \mathbb{R}^{d \times m}$ and tokens in $P$ as $E^p\in \mathbb{R}^{d \times n}$. 

\textbf{Contextual Encoding Layer}. Both passage and question use a shared two-layers BiLSTM as the contextual encoding layer, which projects the lexicon embeddings to contextual embeddings. We  concatenate a pre-trained 600-dimensional CoVe vectors\footnote{https://github.com/salesforce/cove} \cite{mccann2017learned} trained on German-English machine translation dataset, with the aforementioned lexicon embeddings as the final input of the contextual encoding layer, and also with the output of the first contextual encoding layer as the input of its second encoding layer.  To reduce the parameter size, we use a maxout layer \cite{goodfellow2013maxout} at each BiLSTM layer to shrink its dimension. By a concatenation of the outputs of two BiLSTM layers, we obtain $H^q\in \mathbb{R}^{2d \times m}$ as representation of $Q$ and $H^p\in \mathbb{R}^{2d \times n}$ as representation of $P$, where $d$ is the hidden size of the BiLSTM. 

\textbf{Memory Generation Layer}. In the memory generation layer, We construct the working memory, a summary of information from both $Q$ and $P$. 
First, a dot-product attention is adopted like in \cite{vaswani2017attention} to measure the similarity between the tokens in $Q$ and $P$.
Instead of using a scalar to normalize the scores as in \cite{vaswani2017attention}, we use one layer network to transform the contextual information of both $Q$ and $P$:
\vspace{-0.1cm}
\begin{equation}
C=dropout(f_{attention}(\hat{H}^q, \hat{H}^p)) \in \mathbb{R}^{m \times n}\\
\label{eq:align}
\end{equation}
$C$ is an attention matrix. 
Note that $\hat{H^q}$ and $\hat{H^p}$ is transformed from $H^q$ and $H^p$ by one layer neural network $ReLU(W_3x)$, respectively. 
Next, we gather all the information on passages by a simple concatenation of its contextual information $H^p$ and its question-aware representation $H^q \cdot C$:
\begin{equation}
U^p = concat(H^p, H^qC) \in \mathbb{R}^{4d \times n}
\label{eq:gather}
\end{equation}

Typically, a passage may contain hundred of tokens, making it hard to learn the long dependencies within it. Inspired by \cite{lin2017structured}, we apply a self-attended layer to rearrange the information $U^p$ as: 
\begin{equation}
\hat{U}^p = U^p drop_{diag}(f_{attention}(U^p, U^p)).
\label{eq:self}
\end{equation}
In other words, we first obtain an $n \times n$ attention matrix with $U^p$ onto itself, apply dropout, then multiply this matrix with $U^p$ to obtain an updated $\hat{U}^p$.
Instead of using a penalization term as in \cite{lin2017structured}, we dropout the diagonal of the similarity matrix forcing each token in the passage to align to other tokens rather than itself.

At last, the working memory is generated by using another BiLSTM based on all the information gathered:
\begin{equation}
M=BiLSTM([U^p; \hat{U}^p])
\label{eq:mem}
\end{equation}
where the semicolon mark $;$ indicates the vector/matrix concatenation operator.

\textbf{Answer module}.
There is a Chinese proverb that says: ``wisdom of masses exceeds that of any individual."
Unlike other multi-step reasoning models, which only uses a single output either at the last step or some dynamically determined final step, our answer module employs all the outputs of multiple step reasoning. Intuitively, by applying dropout, it avoids a ``step bias problem" (where models places too much emphasis one particular step's predictions) and forces the model to produce good predictions at every individual step. Further, during decoding, we reuse \textit{wisdom of masses} instead of \textit{individual} to achieve a better result.
We call this method ``stochastic prediction dropout" because dropout is being applied to the final predictive distributions. 

Formally, our answer module will compute over $T$ memory steps and output the answer span. 
This module is a memory network and has some similarities to other multi-step reasoning networks: namely, it maintains a state vector, one state per step. 
At the beginning, the initial state $s_0$ is the summary of the $Q$: $s_0=\sum_j \alpha_j H^q_{j}$, where $\alpha_j = \frac{exp(w_4 \cdot H^q_j)}{\sum_{j'}exp(w_4 \cdot H^q_{j'})}$. At time step $t$ in the range of $\{1, 2, ..., T-1\}$, the state is defined by $s_t = GRU(s_{t-1}, x_t)$. 
Here, $x_t$ is computed from the previous state $s_{t-1}$ and memory $M$: $x_t=\sum_j\beta_j M_j$ and $\beta_j = softmax(s_{t-1}W_5M)$. 
Finally, a bilinear function is used to find the begin and end point of answer spans at each reasoning step $t \in \{0,1,\ldots,T-1\}$. \vspace{-0.1cm}
\begin{equation}
P_t^{begin} = softmax(s_tW_6M)
\label{eq:begin}
\end{equation} 
\begin{equation}
P_t^{end} = softmax([s_t; \sum_j P_{t,j}^{begin}M_j]W_7M).
\label{eq:end}
\end{equation}

From a pair of begin and end points, the answer string can be extracted from the passage.
However, rather than output the results (start/end points) from the final step (which is fixed at $T-1$ as in Memory Networks or dynamically determined as in ReasoNet), 
we utilize all of the $T$ outputs by averaging the scores:
\begin{equation}
P^{begin} = avg([P_0^{begin}, P_1^{begin}, ..., P_{T-1}^{begin}])
\label{eq:avg}
\end{equation}
\begin{equation}
P^{end} = avg([P_0^{end}, P_1^{end}, ..., P_{T-1}^{end}])
\label{eq:avg2}
\end{equation} 
Each $P_t^{begin}$ or $P_t^{end}$ is a multinomial distribution over $\{1,\ldots,n\}$, so the average distribution is straightforward to compute. 

During training, we apply stochastic dropout to before the above averaging operation. For example, as illustrated in Figure \ref{fig:answer-module}, we randomly delete several steps' predictions in Equations \ref{eq:avg} and \ref{eq:avg2} so that $P^{begin}$ might be $avg([P_1^{begin}, P_3^{begin}])$ and $P^{end}$ might be $avg([P_0^{end}, P_3^{end}, P_{4}^{end}])$. 
The use of averaged predictions and dropout during training improves robustness. 

Our stochastic prediction dropout is similar in motivation to the dropout introduced by \cite{srivastava2014dropout}. The difference is that theirs is dropout at the intermediate node-level, whereas ours is dropout at the final layer-level.
Dropout at the node-level prevents correlation between features. 
Dropout at the final layer level, where randomness is introduced to the averaging of predictions, prevents our model from relying exclusively on a particular step to generate correct output. 
We used a dropout rate of 0.4 in experiments.

\section{Experiment Setup}
\label{sec:exp}
\vspace{-0.1cm}
\textbf{Dataset}: We evaluate on the Stanford Question Answering Dataset (SQuAD) \cite{rajpurkar2016squad}. This contains about 23K passages and 100K questions. The passages come from approximately 500 Wikipedia articles and the questions and answers are obtained by crowdsourcing. The crowdsourced workers are asked to read a passage (a paragraph), come up with questions, then mark the answer span. All results are on the official development set, unless otherwise noted. 

Two evaluation metrics are used: Exact Match (EM), which measures the percentage of span predictions that matched any one of the ground truth answer exactly, and Macro-averaged F1 score, which measures the average overlap between the prediction and the ground truth answer. 

\textbf{Implementation details}: The spaCy tool\footnote{https://spacy.io} is used to tokenize the both passages and questions, and generate lemma, part-of-speech and named entity tags. We use 2-layer BiLSTM with $d=128$ hidden units for both passage and question encoding. The mini-batch size is set to 32 and Adamax \cite{kingma2014adam} is used as our optimizer. The learning rate is set to 0.002 at first and decreased by half after every 10 epochs. We set the dropout rate for all the hidden units of LSTM, and the answer module output layer to 0.4. To prevent degenerate output, we ensure that at least one step in the answer module is active during training.

\section{Results}
\begin{table*}[ht!]
\centering
\begin{tabular}{ l || c | c}
\hline
Answer Module	&EM&F1\\ \hline\hline
Standard 1-step &75.139 &83.367 \\ \hline
Fixed 5-step with Memory Network (prediction from final step) &75.033 & 83.327  \\ \hline
Fixed 5-step with Memory Network (prediction averaged from all steps) &75.256 & 83.215  \\ \hline
Dynamic steps (max 5) with ReasoNet & 75.355 &  83.360\\ \hline
Stochastic Answer Network (\DR), Fixed 5-step &\textbf{76.235}& {\textbf{84.056}} \\ \hline
\end{tabular}
\caption{\label{tab:model_comp} \textbf{Main results}---Comparison of different answer module architectures. Note that SAN performs best in both Exact Match and F1 metrics.} 
\end{table*}

The main experimental question we would like to answer is whether the stochastic dropout and averaging in the answer module is an effective technique for multi-step reasoning. 
To do so, we fixed all lower layers and compared different architectures for the answer module: 
\begin{enumerate}
\item Standard 1-step: generate prediction from $s_0$, the first initial state.
\item 5-step memory network: this is a memory network fixed at 5 steps. We try two variants: the standard variant outputs result from the final step $s_{T-1}$. The averaged variant outputs results by averaging across all 5 steps, and is like SAN without the stochastic dropout. 
\item ReasoNet\footnote{The ReasoNet here is not an exact re-implementation of \cite{shen2017empirical}. The answer module is the same as \cite{shen2017empirical} but the lower layers are set to be the same as SAN, 5-step memory network, and standard 1-step as described in Figure \ref{fig:model}. We only vary the answer module in our experiments for a fair comparison.}: this answer module dynamically decides the number of steps and outputs results conditioned on the final step.
\item SAN: proposed answer module that uses stochastic dropout and prediction averaging.
\end{enumerate}

The main results in terms of EM and F1 are shown in  Table \ref{tab:model_comp}. 
We observe that SAN achieves 76.235 EM and 84.056 F1, outperforming all other models. 
Standard 1-step model only achieves 75.139 EM and dynamic steps (via ReasoNet) achieves only 75.355 EM. SAN also outperforms a 5-step memory net with averaging, which implies averaging predictions is not the only thing that led to SAN's superior results; 
indeed, stochastic prediction dropout is an effective technique.

\begin{figure}[t]
\centering  
\subfigure[EM comparison on different systems.]{\includegraphics[width=0.8\linewidth]{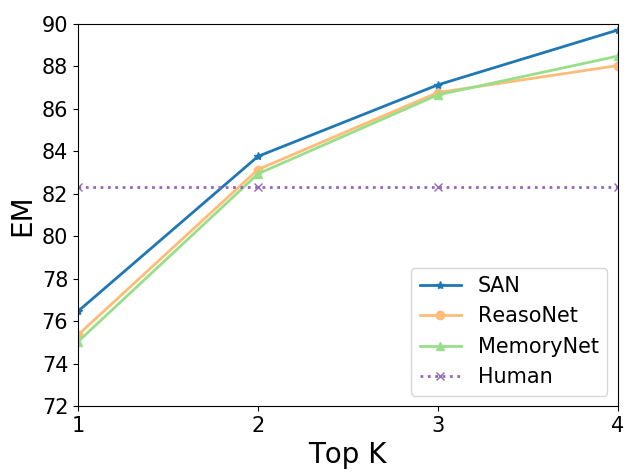}}
\subfigure[F1 score comparison on different systems.]{\includegraphics[width=0.82\linewidth]{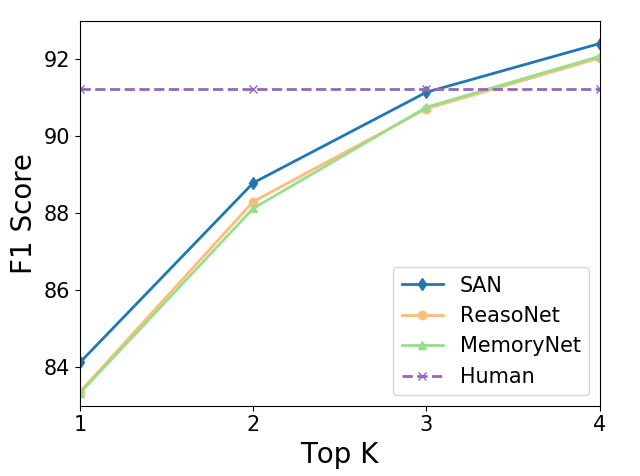}}
\caption{\label{fig:sys_vs_hum} K-Best Oracle results}
\end{figure}

The K-best oracle results is shown in Figure \ref{fig:sys_vs_hum}. 
The K-best spans are computed by ordering the spans according the their probabilities $P^{begin} \times P^{end}$.
We limit K in the range 1 to 4 and then pick the span with the best EM or F1 as oracle. 
SAN also outperforms the other models in terms of K-best oracle scores. Impressively, these models achieve human performance at $K=2$ for EM and $K=3$ for F1.

Finally, we compare our results with other top models in Table \ref{tab:squad_result}. Note that all the results in Table~\ref{tab:squad_result} are taken from the published papers. 
We see that SAN is very competitive in both single and ensemble settings (ranked in second) despite its simplicity. 
Note that the best-performing model \cite{2018arXiv180205365P} used a large-scale language model as an extra contextual embedding, which gave a significant improvement (+4.3\% dev F1). We expect significant improvements if we add this to SAN in future work.

\begin{table*}[t!]
\centering
\begin{tabular}{l || c | c  }
\hline
Ensemble model results:&Dev Set (EM/F1)& Test Set (EM/F1) \\ \hline
\hline
BiDAF + Self Attention + ELMo \cite{2018arXiv180205365P} &-/-&\textbf{81.003}/\textbf{87.432} \\ \hline
\textbf{\DR~(Ensemble model)} &78.619/85.866&79.608/86.496 \\ \hline
AIR-FusionNet \cite{huang2017fusionnet} &-/-&78.978/86.016\\ \hline
DCN+ \cite{xiong2017dcn} &-/-& 78.852/85.996\\ \hline
M-Reader \cite{hu2017mnemonic} &-/-&77.678/84.888 \\ \hline
Conductor-net \cite{liu2017phase}&74.8 / 83.3&76.996/84.630 \\ \hline
r-net \cite{wang2017gated}& 77.7/83.7&76.9/84.0\\ \hline
ReasoNet++ \cite{shen2017empirical} & 75.4/82.9&75.0/82.6 \\\hline
\hline
\multicolumn{3}{l}{\textit{Individual model results:}} \\ \cline{1-3}
BiDAF + Self Attention + ELMo\cite{2018arXiv180205365P} &-/-&\textbf{78.580}/\textbf{85.833} \\ \hline
\textbf{\DR (single model)}  &\textbf{76.235}/\textbf{84.056} &76.828/84.396 \\ \hline
AIR-FusionNet\cite{huang2017fusionnet} &75.3/83.6&75.968/83.900\\ \hline
RaSoR + TR \cite{2017arXiv171203609S} &-/-&75.789/83.261 \\ \hline
DCN+\cite{xiong2017dcn} &74.5/83.1& 75.087/83.081\\ \hline
r-net\cite{wang2017gated}& 72.3/80.6&72.3/80.7 \\ \hline

ReasoNet++\cite{shen2017empirical} &70.8/79.4 &70.6/79.36\\ \hline
BiDAF \cite{seo2016bidirectional} &67.7/77.3&68.0/77.3\\ \hline
 \hline
Human Performance &80.3/90.5&82.3/91.2\\ \hline
\end{tabular}
\caption{\label{tab:squad_result} Test performance on SQuAD. Results are sorted by Test F1.}
\end{table*}

\section{Analysis}
\label{sec:ana}

\subsection{How robust are the results?}

We are interested in whether the proposed model is sensitive to different random initial conditions. 
Table \ref{tab:robustness} shows the development set scores of SAN trained from initialization with different random seeds.
We observe that the SAN results are consistently strong regardless of the 10 different initializations. 
For example, the mean EM score is 76.131 and the lowest EM score is 75.922, both of which still outperform the 75.355 EM of the Dynamic step ReasoNet in Table \ref{tab:model_comp}.\footnote{Note the Dev EM/F1 scores of ReasoNet in Table \ref{tab:model_comp} do not match those of ReasoNet++ in Table \ref{tab:squad_result}. While the answer module is the same architecture, the lower encoding layers are different.}

We are also interested in how sensitive are the results to the number of reasoning steps, which is a fixed hyper-parameter. 
Since we are using dropout, a natural question is whether we can extend the number of steps to an extremely large number. 
Table \ref{tab:step} shows the development set scores for $T=1$ to $T=10$.
We observe that there is a gradual improvement as we increase $T=1$ to $T=5$, but after 5 steps the improvements have saturated. 
In fact, the EM/F1 scores drop slightly, but considering that the random initialization results in Table \ref{tab:robustness} show a standard deviation of 0.142 and a spread of 0.426 (for EM), we believe that the $T=10$ result does not statistically differ from the $T=5$ result. 
In summary, we think it is useful to perform some approximate hyper-parameter tuning for the number of steps, but it is not necessary to find the exact optimal value. 
 
\begin{table}[t]
\centering
\begin{tabular}{@{\hskip1pt} l ||@{\hskip1pt} c |@{\hskip1pt} c |@{\hskip1pt} l |@{\hskip1pt} c |@{\hskip1pt} c @{\hskip1pt}}
\hline
Seed\# &EM&F1 & Seed\# &EM&F1\\ \hline\hline
\textbf{Seed 1} &76.24 &84.06 &Seed 6 &76.23	&83.99\\ \hline
Seed 2 &76.30 &\textbf{84.13} &Seed 7 &\textbf{76.35}	&84.09 \\ \hline
Seed 3 &\textbf{75.92} &83.90 &Seed 8 &76.07 &83.71 \\ \hline
Seed 4 &76.00	&83.95 &Seed 9 &75.93 &\textbf{83.85}\\ \hline
Seed 5 &76.12	&83.99 &Seed 10 &76.15 &84.11\\ \hline
\multicolumn{6}{@{\hskip1pt}r@{\hskip1pt}}{Mean: 76.131, Std. deviation: 0.142 (EM)} \\ 
\multicolumn{6}{@{\hskip1pt}r@{\hskip1pt}}{Mean: 83.977, Std. deviation: 0.126 (F1)} \\
\hline
\end{tabular}
\caption{\label{tab:robustness} 
\textbf{Robustness of \DR (5-step) on different random seeds for initialization}: best and worst scores are boldfaced. Note that our official submit is trained on seed 1.} 
\end{table}
%

\begin{table}[t!]
\centering
\begin{tabular}{ @{\hskip1pt} l || c || c |@{\hskip1pt} c | c | c}
\hline
\DR &EM&F1 &\DR &EM&F1\\ \hline\hline
1 step &\textbf{75.38} &\textbf{83.29} &6 step &75.99&83.72 \\ \hline
2 step &75.43 &83.41 &7 step &76.04 &83.92\\ \hline
3 step &75.89 &83.57 &8 step &76.03 &83.82\\ \hline
4 step &75.92 &83.85 &9 step &75.95 &83.75\\ \hline
5 step &\textbf{76.24}& {\textbf{84.06}} &10 step &76.04 &83.89 \\ \hline
\end{tabular}
\caption{\label{tab:step} \textbf{Effect of number of steps}: best and worst results are boldfaced.} 
\end{table}

Finally, we test SAN on two Adversarial SQuAD datasets, AddSent and AddOneSent \cite{jia2017adversarial}, where the passages contain auto-generated adversarial distracting sentences to fool computer systems that are developed to answer questions about the passages.  
For example, AddSent is constructed by adding sentences that look similar to the question, but do not actually contradict the correct answer. AddOneSent is constructed by appending a random human-approved sentence to the passage.

We evaluate the single SAN model (i.e., the one presented in Table \ref{tab:squad_result}) on both AddSent and AddOneSent. The results in Table~\ref{tab:adversial_data} show that SAN achieves the new state-of-the-art performance and SAN's superior result is mainly attributed to the multi-step answer module, which leads to significant improvement in F1 score over the Standard 1-step answer module, i.e., +1.2 on AddSent and +0.7 on AddOneSent.

 
\begin{table}[t]
\centering
\begin{tabular}{@{\hskip1pt}l @{\hskip1pt}||@{\hskip1pt} c @{\hskip1pt}| @{\hskip1pt}c @{\hskip1pt} }
\hline
Single model:&AddSent& AddOneSent \\ \hline
\hline
LR \cite{rajpurkar2016squad} &23.2 &30.3 \\ \hline
SEDT \cite{liu2017structural} &33.9 & 44.8 \\ \hline
BiDAF \cite{seo2016bidirectional}&34.3 & 45.7 \\ \hline
jNet \cite{zhang2017exploring}&37.9 &47.0\\ \hline
ReasoNet\cite{shen2017empirical} & 39.4& 50.3 \\ \hline
RaSoR\cite{lee2016learning} &39.5 &49.5\\ \hline
Mnemonic\cite{hu2017mnemonic} & \textbf{46.6}& 56.0\\\hline \
QANet\cite{yu18qanet} &45.2 &55.7 \\ \hline \hline
Standard 1-step in Table~\ref{tab:model_comp} &45.4& 55.8\\ \hline
SAN &\textbf{46.6}& \textbf{56.5}\\ \hline
\end{tabular}
\caption{\label{tab:adversial_data} Test performance on the adversarial SQuAD dataset in F1 score.}
\end{table}

\subsection{Is it possible to use different numbers of steps in test vs. train?} 
For practical deployment scenarios, prediction speed at test time is an important criterion.
Therefore, one question is whether SAN can train with, e.g. $T=5$ steps but test with $T=1$ steps. 
Table \ref{tab:comp_pre_step} shows the results of a SAN trained on $T=5$ steps, but tested with different number of steps. 
As expected, the results are best when $T$ matches during training and test; however, it is important to note that small numbers of steps $T=1$ and $T=2$ nevertheless achieve strong results.
For example, prediction at $T=1$ achieves 75.58, which outperforms a standard 1-step model (75.14 EM) as in Table~\ref{tab:model_comp} that has approximate equivalent prediction time. 

\begin{table}[t!]
\centering
\begin{tabular}{@{\hskip1pt} l ||@{\hskip1pt} c |@{\hskip1pt} c |@{\hskip1pt} l |@{\hskip1pt} c |@{\hskip1pt} c}
\hline
$T=$	&EM&F1 &$T=$	&EM&F1\\ \hline\hline
1 &75.58 &83.86 &4 &76.12 &83.98\\ \hline
2 &75.85 &83.90 &5 &76.24 &84.06 \\ \hline
3 &75.98 &83.95 &10 &75.89 &83.88 \\ \hline
\end{tabular}
\caption{\label{tab:comp_pre_step} 
\textbf{Prediction on different steps $T$.} Note that the SAN model is trained using 5 steps.} 
\end{table}

\subsection{How does the training time compare?}
\begin{figure}[t!]
\centering  
\subfigure[EM]{\includegraphics[width=0.9\linewidth]{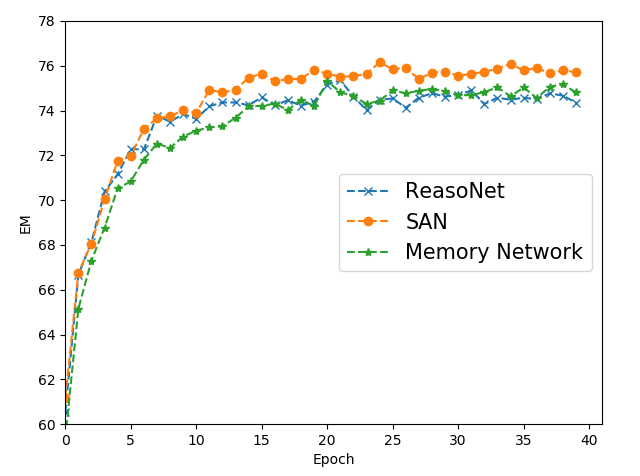}}
\subfigure[F1]{\includegraphics[width=0.9\linewidth]{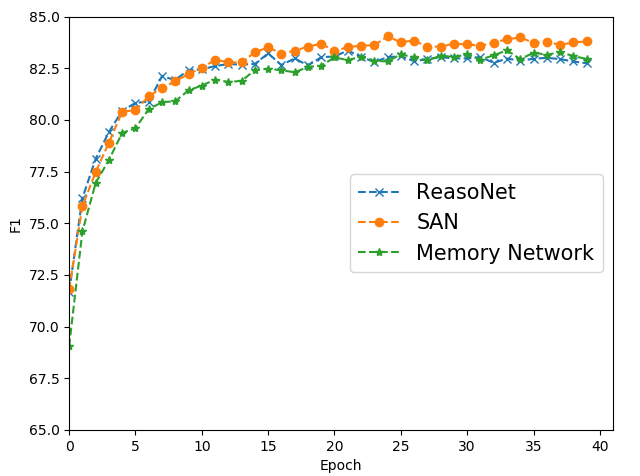}}
\caption{\label{fig:dev_curve} Learning curve measured on Dev set.}
\end{figure}

The average training time per epoch is comparable: our implementation running on a GTX Titan X
is 22 minutes for 5-step memory net, 30 minutes for ReasoNet, and 24 minutes for SAN. 
The learning curve is shown in Figure \ref{fig:dev_curve}.
We observe that all systems improve at approximately the same rate up to 10 or 15 epochs. 
However, SAN continues to improve afterwards as other models start to saturate. 
This observation is consistent with previous works using dropout \cite{srivastava2014dropout}. 
We believe that while training time per epoch is similar between SAN and other models, it is recommended to train SAN for more epochs in order to achieve gains in EM/F1. 

\subsection{How does SAN perform by question type?}
\begin{figure}[t!]
\centering
\includegraphics[scale=0.31]{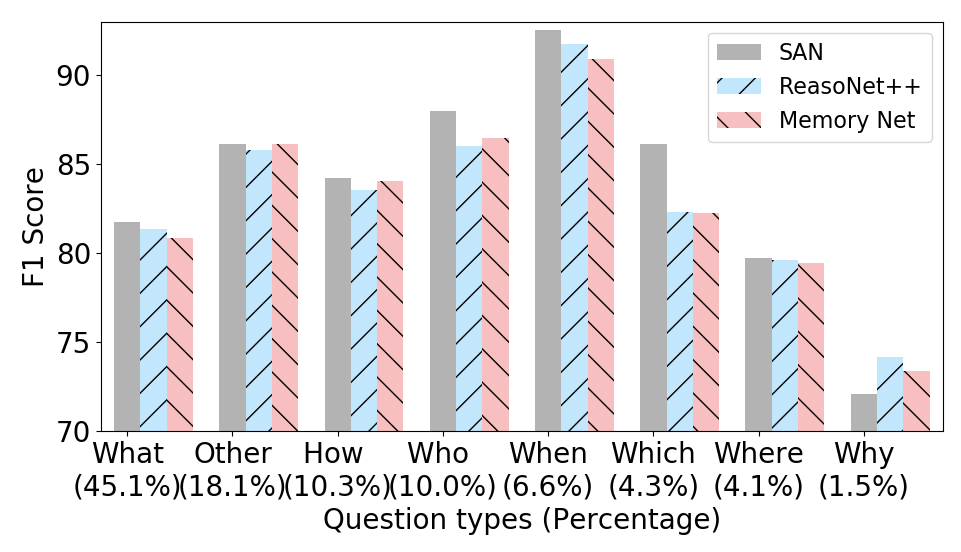}
\caption{\label{fig:dev_f1}Score breakdown by question type.}
\end{figure}

To see whether SAN performs well on a particular type of question, we divided the development set by questions type based on their respective Wh-word, such as ``who" and ``where". 
The score breakdown by F1 is shown in Figure \ref{fig:dev_f1}. 
We observe that SAN seems to outperform other models uniformly across all types. 
The only exception is the Why questions, but there is too little data to derive strong conclusions. 

\subsection{Experiments results on MS MARCO}
MS MARCO \cite{nguyen2016ms} is a large scale real-word RC dataset which contains 100,100 (100K) queries collected from anonymized user logs from the Bing search engine. The characteristic of MS MARCO is that  all the questions are real user queries and passages are extracted from real web documents. For each query, approximate 10 passages are extracted from public web documents. The answers are generated by humans. The data is partitioned into a 82,430 training, a 10,047 development and 9,650 test tuples. The evaluation metrics are BLEU\cite{papineni2002bleu} and ROUGE-L \cite{lin2004rouge} due to its free-form text answer style. To apply the same RC model, we search for a span in MS MARCO's passages that maximizes the ROUGE-L score with the raw free-form answer. It has an upper bound of 93.45 BLEU and 93.82 ROUGE-L on the development set. 

The MS MARCO dataset contains multiple passages per query. Our model as shown in Figure~\ref{fig:model} is developed to generate answer from a single passage. Thus, we need to extend it to handle multiple passages. Following \cite{shen2017empirical}, we take two steps to generate an answer to a query $Q$ from $J$ passages, $P^1, ..., P^J$. First, we run SAN on every ($P^j, Q$) pair, generating $J$ candidate answer spans, one from each passage. Then, we multiply the SAN score of each candidate answer span with its relevance score $r(P^j, Q)$ assigned by a passage ranker, and output the span with the maximum score as the answer. In our experiments, we use the passage ranker described in \cite{liu2018stochastic}\footnote{It is the same model structure as \cite{liu2018stochastic} by using softmax over all candidate passages. A simple baseline, TF-IDF, obtains 20.1 p@1 on MS MARCO development.}. The ranker is trained on the same MS MARCO training data, and achieves 37.1 p@1 on the development set.



\begin{table}[t!]
\centering
\begin{tabular}{@{\hskip1pt} l @{\hskip1pt} ||@{\hskip1pt} c @{\hskip1pt} | @{\hskip1pt} c }
\hline
$Single Model$	& ROUGE & BLEU \\ \hline\hline
ReasoNet++\cite{shen2017empirical} &38.01 & 38.62\\ \hline
V-Net\cite{multi-passage-mrc2018} & 45.65 & - \\ \hline \hline
Standard 1-step in Table~\ref{tab:model_comp}  &42.30 &42.39 \\ \hline 
SAN &\textbf{46.14} & \textbf{43.85} \\ \hline
\end{tabular}
\caption{\label{tab:comp_marco} 
\textbf{MS MARCO devset results}.} 
\end{table}


The results in Table~\ref{tab:comp_marco} show that SAN outperforms V-Net \cite{multi-passage-mrc2018} and becomes the new state of the art\footnote{The official evaluation on MS MARCO on test is closed, thus here we only report the results on the development set.}.

\section{Related Work}
The recent big progress on MRC is largely due to the availability of the large-scale datasets \cite{rajpurkar2016squad, nguyen2016ms, richardson13mctest, hill2015goldilocks}, since it is possible to train large end-to-end neural network models. 
In spite of the variety of model structures and attenion types \cite{bahdanau2014neural, chen2016thorough, xiong2016dynamic, seo2016bidirectional, shen2017empirical, wang2017gated}, a typical neural network MRC model first maps the symbolic representation of the documents and questions into a neural space, then search answers on top of it. We categorize these models into two groups based on the difference of the answer module: single-step and multi-step reasoning.
The key difference between the two is what strategies are applied to search the final answers in the neural space.

A single-step model matches the question and document only once and produce the final answers. It is simple yet efficient and can be trained using the classical back-propagation algorithm, thus it is adopted by most systems \cite{chen2016thorough, seo2016bidirectional, wang2017gated, liu2017phase, chen2017reading, weissenborn2017fastqa, hu2017mnemonic}. However, since humans often solve question answering tasks by re-reading and re-digesting the document multiple times before reaching the final answers (this may be based on the complexity of the questions/documents), it is natural to devise an iterative way to find answers as multi-step reasoning.

Pioneered by \cite{hill2015goldilocks,dhingra2016gated,sordoni2016iterative,kumar15askme}, who used a predetermined fixed number of reasoning steps, Shen et al \shortcite{shen2016reasonet, shen2017empirical} showed that multi-step reasoning outperforms single-step ones and dynamic multi-step reasoning further outperforms the fixed multi-step ones on two distinct MRC datasets (SQuAD and MS MARCO). But these models have to be trained using reinforcement learning methods, e.g., policy gradient, which are tricky to implement due to the instability issue.
Our model is different in that we fix the number of reasoning steps, but perform stochastic dropout to prevent step bias. Further, our model can also be trained by using the back-propagation algorithm, which is simple and yet efficient.

\section{Conclusion}
We introduce Stochastic Answer Networks (SAN), a simple yet robust model for machine reading comprehension. The use of stochastic dropout in training and averaging in test at the answer module leads to robust improvements on SQuAD, outperforming both fixed step memory networks and dynamic step ReasoNet. We further empirically analyze the properties of SAN in detail.
The model achieves results competitive with the state-of-the-art on the SQuAD leaderboard, as well as on the Adversarial SQuAD and MS MARCO datasets.
Due to the strong connection between the proposed model with memory networks and ReasoNet, we would like to delve into the theoretical link between these models and its training algorithms. Further, we also would like to explore SAN on other tasks, such as text classification and natural language inference for its generalization in the future.

\section*{Acknowledgments}
We thank Pengcheng He, Yu Wang and Xinying Song for help to set up dockers. We also thank Pranav Samir Rajpurkar for help on SQuAD evaluations, and the anonymous reviewers for valuable discussions
and comments.
%
\bibliography{naacl-deep-reader}
\bibliographystyle{acl_natbib}

\end{document}